\documentclass[10pt,twocolumn,letterpaper]{article}

\usepackage{cvpr}              

\usepackage{times}
\usepackage{epsfig}
\usepackage{graphicx}
\usepackage{amsmath}
\usepackage{amssymb}
\usepackage{amssymb}
\usepackage{bbm}
\DeclareMathAlphabet\mathbfcal{OMS}{cmsy}{b}{n}

\usepackage{lineno}
\usepackage{makecell}
\usepackage{float}

\usepackage{lipsum}
\usepackage{stfloats}
\usepackage{multicol}
\usepackage{multirow}
\usepackage{bm}
\usepackage{etoolbox}

\usepackage{array}
\usepackage{tabulary}
\usepackage[table]{xcolor}
\usepackage{paralist}
\usepackage{booktabs}
\usepackage{adjustbox}

\usepackage[acronym]{glossaries}
\newacronym[
    shortplural={GPUs},
    longplural={Graphics Processing Units}
] {gpu}{GPU}{Graphics Processing Unit}

\newacronym[] {cnn}{CNN}{Convolutional Neural Network}

\newacronym[] {dacs}{DACS}{Domain Adaptation via Cross-domain Mixed Sampling}

\newacronym[] {sota}{SOTA}{state-of-the-art}

\newacronym[] {lr}{LR}{learning rate}

\newacronym[] {mldg}{MLDG}{meta-learning for domain generalization}

\newacronym[] {lawinaspp}{LawinASPP}{Large Window Attention Spatial Pyramid Pooling}

\newacronym[] {pov}{POV}{point-of-view}

\newacronym[] {bvip}{BVIP}{blind and visually impaired people}

\newacronym[] {hog}{HOG}{histogram of oriented gradients}

\newacronym[] {svm}{SVM}{support vector machine}

\newacronym[] {fps}{FPS}{frame per second}
\newacronym[] {ocda}{OCDA}{Open compound domain adaptation}

\newacronym[] {da}{DA}{Domain Adaptation}

\newacronym[] {dg}{DG}{Domain Generalization}
\newacronym[] {iou}{IoU}{Intersection over Union}
\newacronym[] {miou}{mIoU}{Mean Intersection over Union}

\usepackage{caption}
\usepackage{subcaption}

\usepackage[pagebackref=true,breaklinks=true,colorlinks,bookmarks=true,bookmarksopen=true]{hyperref}
\hypersetup{colorlinks, citecolor=blue}

\definecolor{lightred}{HTML}{F19C99}

\usepackage[capitalize]{cleveref}
\crefname{section}{Sec.}{Secs.}
\Crefname{section}{Section}{Sections}
\Crefname{table}{Table}{Tables}
\crefname{table}{Tab.}{Tabs.}

\makeatletter
\renewcommand*{\@fnsymbol}[1]{\ensuremath{\ifcase#1\or *\or \dagger\or \ddagger\or
    \mathsection\or \mathparagraph\or \|\or **\or \dagger\dagger
    \or \ddagger\ddagger \else\@ctrerr\fi}}
\makeatother

\begin{document}

\title{Towards Robust Semantic Segmentation of Accident Scenes via Multi-Source Mixed Sampling and Meta-Learning}

\author{Xinyu Luo\thanks{The first two authors contribute equally to this work.}\, ,
~Jiaming Zhang$^{*}$,
~Kailun Yang\thanks{Corresponding author (e-mail: {\tt kailun.yang@kit.edu}).}\, ,
~Alina Roitberg,
~Kunyu Peng,
~Rainer Stiefelhagen\\
\normalsize
CV:HCI Lab, Karlsruhe Institute of Technology
}

\maketitle

\begin{abstract}
    Autonomous vehicles utilize urban scene segmentation to understand the real world like a human and react accordingly. Semantic segmentation of \emph{normal} scenes has experienced a remarkable rise in accuracy on conventional benchmarks. However, a significant portion of real-life accidents features \emph{abnormal} scenes, such as those with object deformations, overturns, and unexpected traffic behaviors. Since even small mis-segmentation of driving scenes can lead to serious threats to human lives, the robustness of such models in accident scenarios is an extremely important factor in ensuring safety of intelligent transportation systems.

In this paper, we propose a \emph{Multi-source Meta-learning Unsupervised Domain Adaptation (MMUDA)} framework, to improve the generalization of segmentation transformers to extreme accident scenes. In MMUDA, we make use of \emph{Multi-Domain Mixed Sampling} to augment the images of multiple-source domains (normal scenes) with the target data appearances (abnormal scenes). To train our model, we intertwine and study a meta-learning strategy in the multi-source setting for robustifying the segmentation results. We further enhance the segmentation backbone (SegFormer) with a \emph{HybridASPP} decoder design, featuring large window attention spatial pyramid pooling and strip pooling, to efficiently aggregate long-range contextual dependencies. Our approach achieves a \acrshort{miou} score of $46.97\%$ on the DADA-seg benchmark, surpassing the previous state-of-the-art model by more than $7.50\%$.\footnote{Code will be made publicly available at \url{https://github.com/xinyu-laura/MMUDA}.}
\end{abstract}

\section{Introduction}
\label{sec:introduction}

\begin{figure}
    \centering
    \subfloat[Image]{\includegraphics[width=0.25\columnwidth]{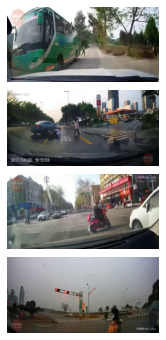}\label{fig:1_orig}} \hspace*{-0.5em}
    \subfloat[GT]{\includegraphics[width=0.25\columnwidth]{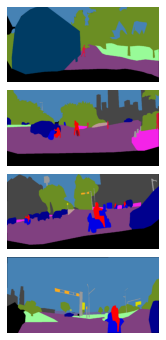}\label{fig:1_gt}} \hspace*{-0.5em}
    \subfloat[ResNet]{\includegraphics[width=0.25\columnwidth]{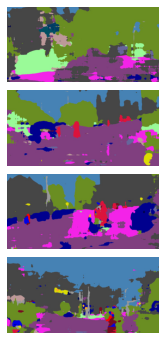}\label{fig:1_pred}} \hspace*{-0.5em}
    \subfloat[Ours]{\includegraphics[width=0.25\columnwidth]{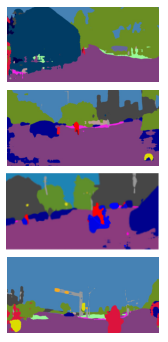}\label{fig:1_ours}} \hspace*{-0.5em}
    \vskip-2ex
    \caption{\textbf{Semantic segmentation of accident scenes.} Compared to the source-only model (\eg, a ResNet), our model generalizes better in the abnormal cases. From top to bottom are dash-crossing pedestrians, overturned motorcycles, collisions, and motion blur.}
    \vskip-3ex
    \label{fig:1}
\end{figure}

With the rapid development of computer vision algorithms in Intelligent Transportation Systems (ITS), road safety for Intelligent Vehicles (IV) has gradually become one of the most concerning issues in this community.
The Advanced Driver Assistance Systems (ADAS) are required to correctly handle both, \textit{normal} driving scenes, which are addressed by most of the published datasets, and \textit{abnormal} situations (\ie, edge cases) that may unexpectedly appear in real-world scenes.
Fueled by rapid improvements in general semantic segmentation research, great progress has been made in the field of autonomous driving~\cite{orsic2019defense,romera2017erfnet,peng2021mass} in recent years.
However, these segmentation models are mainly designed for normal driving scenes, while real-life accidents often encounter abnormalities and critical situations, such as overturned vehicles in front of the ego-vehicle or distorted shots caused by motion blur. 
Several examples of such abnormal cases taken from accident scenes are presented in Fig.~\ref{fig:1}.
If a standard semantic segmentation model, which does not see any abnormal samples during training, is deployed in real world, it can hardly obtain correct results when encountering an unusual accident- or near-accident scene, resulting in a failure of driving assistance.
The large domain gap between the normal- and accident scenes negatively impacts segmentation performance~\cite{zhang2020issafe}, greatly limiting applications of autonomous driving in practice.

Despite its high relevance for applications, only a few works address the task of accident scenes segmentation, which aims to adapt models trained in normal scenarios to the abnormal ones. 
\cite{zhang2020issafe,zhang2021exploring} introduced DADA-seg -- a new traffic accidents dataset covering labelled and unlabelled images from real-world abnormal driving scenes.
The event-aware ISSAFE architecture~\cite{zhang2020issafe} was proposed to fuse RGB images and event data and therefore to capture the dynamic context. 
The Trans4Trans model~\cite{zhang2021trans4trans} leveraged transformer-based encoder and decoder and was simply transferred, without any adaptation design, from multiple source datasets, \ie, Cityscapes~\cite{cordts2016cityscapes} and ACDC~\cite{sakaridis2021acdc}. 
However, mixing data only in a dataset-wise manner is limited in terms of balancing the data distribution and further data diversification remains an important milestone.
To the best of our knowledge, only the single-source Unsupervised Domain Adaptation~(UDA)~\cite{zhang2020issafe} and the multi-source transfer-learning method~\cite{zhang2021trans4trans} from normal- to abnormal domain were investigated.
\emph{Multi-source normal-to-abnormal UDA, yet, remains unexplored.}
We believe, that leveraging multiple data origins has strong potential for robustifying accident scene segmentation, as such extreme scenarios often contain diverse abnormal factors and composite scenes, that could be better addressed by exploiting rich ontologies covered in diverse sources.

To improve the robustness of semantic segmentation in accident scenarios, we propose a novel \emph{Multi-source Meta-learning UDA} framework to help the transformer models in generalization to the unusual target scenes (\emph{MMUDA} for short).
Our framework learns from the label-rich datasets of conventional driving scenes (\textit{source}), and then automatically adapts to the target domain of abnormal accident scenes with only unlabelled training data (\textit{target}).
To effectively learn from the entire unlabelled target domain dataset, we put forward a \emph{Multi-Domain Mixed Sampling (MDMS)} strategy, which is inspired by the cross-domain mixing approach in DACS~\cite{tranheden2021dacs} and augments the training samples of multiple source domains.
Two major differences compared to the \emph{normal-to-normal} DACS are that \textit{i)} we adapt the single-source method to a multi-source setup, and \textit{ii)} we further investigate the multi-source mixing technique in our \emph{normal-to-abnormal} setting.
More precisely, in the case of single-domain mixed sampling, the augmented sample is formed by mixing the source normal image and the target abnormal image.
In the case of the multi-domain mixing, some marks from each source domain image are extracted and then pasted onto the target domain image.
The pseudo-labels for the augmented image are mixed by the source ground-truth labels and the target pseudo labels.

In the training phase, we use the \acrfull{mldg} strategy~\cite{li2018learning}, which was adapted in \cite{zhang2021generalizable} to model the domain transfer problem with an episodic training paradigm, leading to superior performance in image classification.
\acrshort{mldg} can be viewed as a regularization mechanism that prevents the model from overfitting.
Different from the original MLDG, our MMUDA framework performs meta-learning across multiple source domains and jointly with the target domain, after which we apply it to the normal-to-abnormal UDA setting.
In addition, we enhance the segmentation backbone (SegFormer) with a \emph{HybridASPP} decoder design, which leverages large window attention spatial pyramid pooling~\cite{yan2022lawin} and strip pooling~\cite{hou2020strip} with a long but narrow kernel. The HybridASPP decoder replaces the vanilla MLP-based decoder of SegFormer, and this helps to efficiently extract large regions of global context and long-range dependencies.
Comprehensive experiments demonstrate the effectiveness of our proposed methods.
On the challenging DADA-seg benchmark~\cite{zhang2020issafe}, our approach achieves a mIoU score of $46.97\%$, surpassing the previous state-of-the-art transformer model~\cite{zhang2021trans4trans} by more than ${+}7.50\%$.

Our contributions are summarized as follows:

\begin{compactitem}
     \item We propose a novel \emph{Multi-source Meta-learning UDA (MMUDA)} framework for better adaptation from  multi-source domains of normal driving scenes to the domain of abnormal accident scenes.
    \item We develop a \emph{Multi-Domain Mixed Sampling~(MDMS)} approach to augment the training data from multiple labelled source domains with data appearances from the unlabelled target domain data.
    \item We employ meta-learning and analyze its effects under different combinations of multiple source datasets.
    \item We introduce an enhanced \emph{HybridASPP} to replace the vanilla MLP-based decoder of SegFormer, which makes the framework more efficient and effective.
    
\end{compactitem}

\section{Related Work}
\label{sec:related_work}

\noindent\textbf{Semantic segmentation.}
Semantic segmentation has experienced a great breakthrough since the emergence of FCN~\cite{long2015fully} classifying pixels end-to-end.
Subsequent networks, \eg, \cite{chen2017deeplab,Zhao_2017_CVPR,fu2019dual,hou2020strip,yuan2020object_contextual_representations} improved FCN in different aspects, significantly pushing segmentation performance, but also raising computational cost.
To alleviate this issue, compact segmentation models~\cite{orsic2019defense,poudel2019fast,romera2017erfnet} are designed to hold a better accuracy-efficiency trade-off.
Disentangled non-local blocks~\cite{yang2021capturing,yin2020dnl,zhu2019asymmetric} have also been explored to efficiently collect omni-range dependencies.
Recently, the semantic segmentation field has been driven by the newly emerged and highly effective transformer-based architectures~\cite{zheng2021rethinking,strudel2021segmenter,cheng2021maskformer,wang2022dual,yan2022lawin}.
Compared to FCN-based approaches, such models are able to handle long-range dependencies by design and have quickly climbed to the top of segmentation benchmarks.
Furthermore, MLP-like architectures~\cite{chen2021cyclemlp,hou2022vision_permutator,tolstikhin2021mlp,yu2021metaformer,zhang2022bending} alternate token- and channel-mixing to enhance global reasoning.
The recently proposed SegFormer architecture~\cite{xie2021segformer} leverages a hierarchical transformer encoder with a lightweight All-MLP decoder, generating powerful representations without complex and computationally demanding modules.
In this work, we build on SegFormer and introduce multiple building blocks for its improvement specifically for accident scene segmentation.

\noindent\textbf{Domain adaptation and generalization.}
While current semantic segmentation models achieve excellent performance on standard benchmarks, the performance drops sharply if the training and test images come from different domains.
To counter this effect, multiple methods based on \acrfull{da}~\cite{He_2021_CVPR,romera2019bridging,sun2019see,Zhang_2021_CVPR,zhang2017curriculum} were proposed for automatic adjustment to adverse conditions.
In~\cite{saleh2018effective}, effective usage of synthetic data was explored to better handle domain shifts, with results indicating that the foreground objects should rather be addressed in a detection-based manner.
\acrfull{dg} is more challenging than \acrshort{da} since \acrshort{dg} methods can only access source domain data for training.
Target images during the training process cannot be used or observed (while classical \acrshort{da} allows access to unlabelled target domain data).
Currently, most \acrshort{dg} methods focus on image classification, and only a few~\cite{zhang2021generalizable,Choi_2021_CVPR,pan2018two_at_once} are developed to solve semantic segmentation tasks.
Another recent group of approaches \cite{domain_randomization,Huang_2021_CVPR} raised the technique of domain randomization to improve \acrshort{dg}.
Moreover, open compound domain adaptation approaches~\cite{gong2021cluster,liu2020open,park2020discover} have been developed to adapt to a group of unknown heterogeneous domains like  scenes in adverse environmental conditions. 
In this work, we also develop a domain transfer system based on meta-learning and design a multi-source mixed sampling method for enhancing semantic segmentation in accident scenarios.

\noindent\textbf{Meta learning.}
Meta-learning aims at figuring out \textit{how to learn} and has been effectively applied in different fields, with Model-Agnostic Meta-Learning (MAML)~\cite{finn2017model} and HyperNetworks~\cite{ha2017hypernetworks} being popular approches for image classification.
MAML simulates the domain gap between train and test domains by synthesizing virtual testing domains within each mini-batch during training.
As of late, meta-learning has likewise been effectively applied to domain adaptation and domain generalization.
Meta-online~\cite{li2020online} introduces a strategy to improve the results by learning the underlying circumstances (\eg, model boundary) of the existing  domain adaptation strategies.
MLDG~\cite{li2018learning} follows the MAML~\cite{finn2017model} system and has been effectively used in the DG task.
Similar to MAML, the DG task expects that the learned models in seen domains are able to generalize well to novel unseen domains.
As a consequence, meta learning has been recently explored for domain adaptation and generalization~\cite{balaji2018metareg,chen2019blending,dou2019domain_generalization,gong2021cluster,zhang2021generalizable}.
In this work, we leverage meta learning to better make use of multi-source data in order to boost generalization of semantic segmentation models in extreme accident scenes.

\begin{figure*}[!t]
    \centering
    \includegraphics[width=0.9\textwidth]{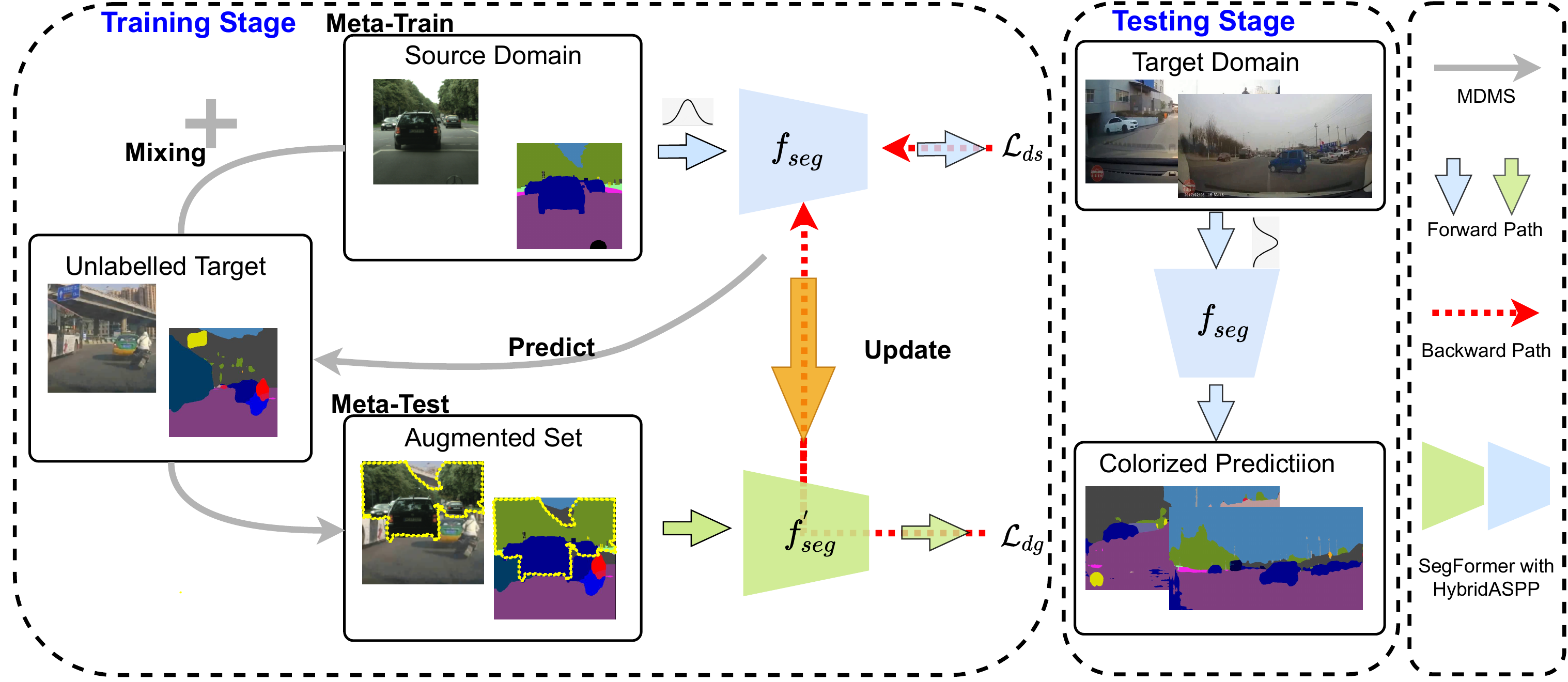}
    \vskip-1ex
    \caption{\textbf{Overview of the proposed \emph{Multi-source Meta-learning UDA (MMUDA)} framework.} It includes \emph{Multi-domain Mixed Sampling (MDMS)} and meta-learning with segmentation transformers. Given multiple source (normal) domains, the model fine-tuned by meta-training and meta-testing across various source domains with rich ontologies, can generalize well in the target (abnormal) domain.}
    \label{fig:framework}
    \vskip-3ex
\end{figure*}

\noindent\textbf{Mixing.}
Mixing is a kind of augmentation technique and has successfully been used for both classification and semantic segmentation as suggested in~\cite{zhang2017mixup}.
Mixing has been further developed in the class mixing algorithm~\cite{olsson2021classmix}, where the masks used for mixing are dynamically created based on the predictions of the network.
In DACS~\cite{tranheden2021dacs}, the concept of self-training is extended and combined with ClassMix.
Their proposed method fine-tunes models with mixed labels by combining ground-truth annotations from the source domain and pseudo-labels from the target domain.
Further, context-aware mixing~\cite{zhou2021context} and bi-mix~\cite{yang2021bi} methods are developed to better guide the domain mixup and bridge the distribution gap.
In this work, we assemble a multi-source mixed sampling method within our system for robust semantic understanding of accident scenes.

\section{Methodology}
\label{sec:methodology}

In this Section, we describe our \emph{Multi-source Meta-learning UDA (MMUDA)} framework in detail. 
First, we explain the proposed \emph{Multi-Domain Mixed Sampling (MDMS)} in Sec.~\ref{sec3:1}.
Then, the meta-learning strategy for multi-source UDA in Sec.~\ref{sec3:2} is explained in detail.
Finally, Sec. ~\ref{sec3:3} provides an overview of the proposed architectural changes and the enhanced \emph{HybridASPP} decoder design.

\subsection{MDMS: Multi-Domain Mixed Sampling}
\label{sec3:1}
Our proposed method builds upon the idea of domain adaptation via cross-domain mixed sampling (DACS)~\cite{tranheden2021dacs}. Unlike DACS, which has only one source domain, our augmented samples are created by mixing pixels from the target domain image with pixels from each source domain image. Before being mixed, the unlabelled target domain images first need to be run through the model to generate pseudo-labels for them.
Then, half of the classes in the image from source domain are randomly selected, and the pixels of the corresponding classes are cut from the source domain image and pasted onto the target domain image.
For labels, the pseudo-labels of unlabelled target image are mixed with the corresponding ground-truth labels of the source domain image in the same way as the mixed images.
The above mixing approach is applied to each source domain. For brevity, the mixed sampling process will be described with only one source domain below.

A source domain is defined by a set of image and label pairs $\{(X_{S}^i, Y_{S}^i)\}^{N_S}{\in} \mathcal{D}_{S}$, where $X_{S}^i{\in}\mathbb{R}^{H{\times}W {\times}3}$ is the image, $Y_{S}^i{\in} \mathbb{R}^{H{\times}W {\times}C}$ is the $C$-class label, and $N_S$ is the number of samples in the source domain $\mathcal{D}_{S}$.
From the target domain $\mathcal{D}_{T}$ with a number of $N_T{=}N_L{+}N_U$ samples, the $N_U$ unlabelled image and pseudo label pairs $\{(X_{T}^i, \hat{Y}_{T}^i)\}^{N_U}{\in} \mathcal{D}_{T}$ are selected for the mixing approach, where $\hat{Y}_{T}$ is generated by the segmentation transformer $f_{seg}$ in Fig.~\ref{fig:framework}. The labelled $N_L$ images in the target domain are only used in the testing stage.
In the augmented set $\mathcal{D}_{M}$ with the same $N_U$ samples, an augmented image $X_M$ is generated by mixing a source image $X_S$ and a target image $X_T$, and the pseudo label $\hat{Y}_M$ by combining the corresponding ground-truth label $Y_S$ and the pseudo label $\hat{Y}_T$. However, in our case, there is not one but $K$ source domains, thus the augmented set is finally created as $\{(X_{M_k}^i, \hat{Y}_{M_k}^i)\}^{N_U}{\in}\mathcal{D}_M$.

These new samples are then used to train the segmentation transformer model $f_{seg}^{'}$.
The process of MDMS is presented as \textcolor{gray}{gray arrows} in Fig.~\ref{fig:framework}, where an example of mixing image-label pairs from source- and target domain is shown. Pixels cut from the source are marked in yellow dotted boxes.
In this way, we obtain a set of augmented samples from multiple source domains, which are then passed to the subsequent meta-learning-based network together with the original source domain images.

\subsection{Meta-Learning in UDA}
\label{sec3:2}
During the training phase, we adapt MLDG~\cite{li2018learning} (\ie, meta-learning for domain generalization) to train our model.
Different from MLDG, we perform meta-learning across multiple source domains and together with the target domain, intertwined via mixing sampling to obtain augmented images which instills the knowledge from the target scenes.
Besides, instead of addressing image classification, we apply it to the normal-to-abnormal UDA setting for robustifying dense accident scene segmentation.

As shown in the training stage of our framework in Fig.~\ref{fig:framework}, we use the images from sources for meta-training while the augmented images produced via MDMS are used for meta-testing.
The cross-entropy is selected as the loss function $\mathcal{L}$ for our semantic segmentation task.
First, the domain-specific loss $\mathcal{L}_{ds}$ is computed from the meta-training data though the network $f_{seg}$.
The gradient $\nabla \mathcal{L}_{ds}$ is then used to update a new network $f_{seg}^{'}$, \ie, the green block in Fig.~\ref{fig:framework}, which has the same backbone as the blue one. 
As we want the model to adapt well in the unseen target domain, the domain adaptation loss $\mathcal{L}_{da}$ is calculated from $f_{seg}^{'}$ with the updated parameters using the meta-test data.
Finally, we employ the total loss $\mathcal{L}_{total} {=} \mathcal{L}_{da} {+} \alpha\mathcal{L}_{ds}$ to update the original $f_{seg}$, so that the network is optimized to perform well in the source and target domains. During the update for $f_{seg}^{'}$, we use the inner learning rate $\eta$, while updating the original network $f_{seg}$ after a meta-train and subsequent meta-test process, the outer learning rate $\gamma$ is used. 
The parameters can then be updated with SGD (\ie, Stochastic Gradient Descent).

In our segmentation task, the statistics in the source domain (\ie, \textit{the normal driving scenes}) are different from the target domain of traffic accident scenes.
Therefore, normalizing the test data with the accumulated mean and variance during the training phase can be problematic. Considering this fact, we adopt target-specific normalization introduced by \cite{zhang2021generalizable} to normalize features directly using statistics from the target domain in the testing stage.
In the experiments, we further investigate different combinations of datasets for meta-learning, aiming for an optimal path to follow towards robust accident scene understanding.

\begin{figure*}[!t]
    \centering
    \includegraphics[width=0.9\textwidth]{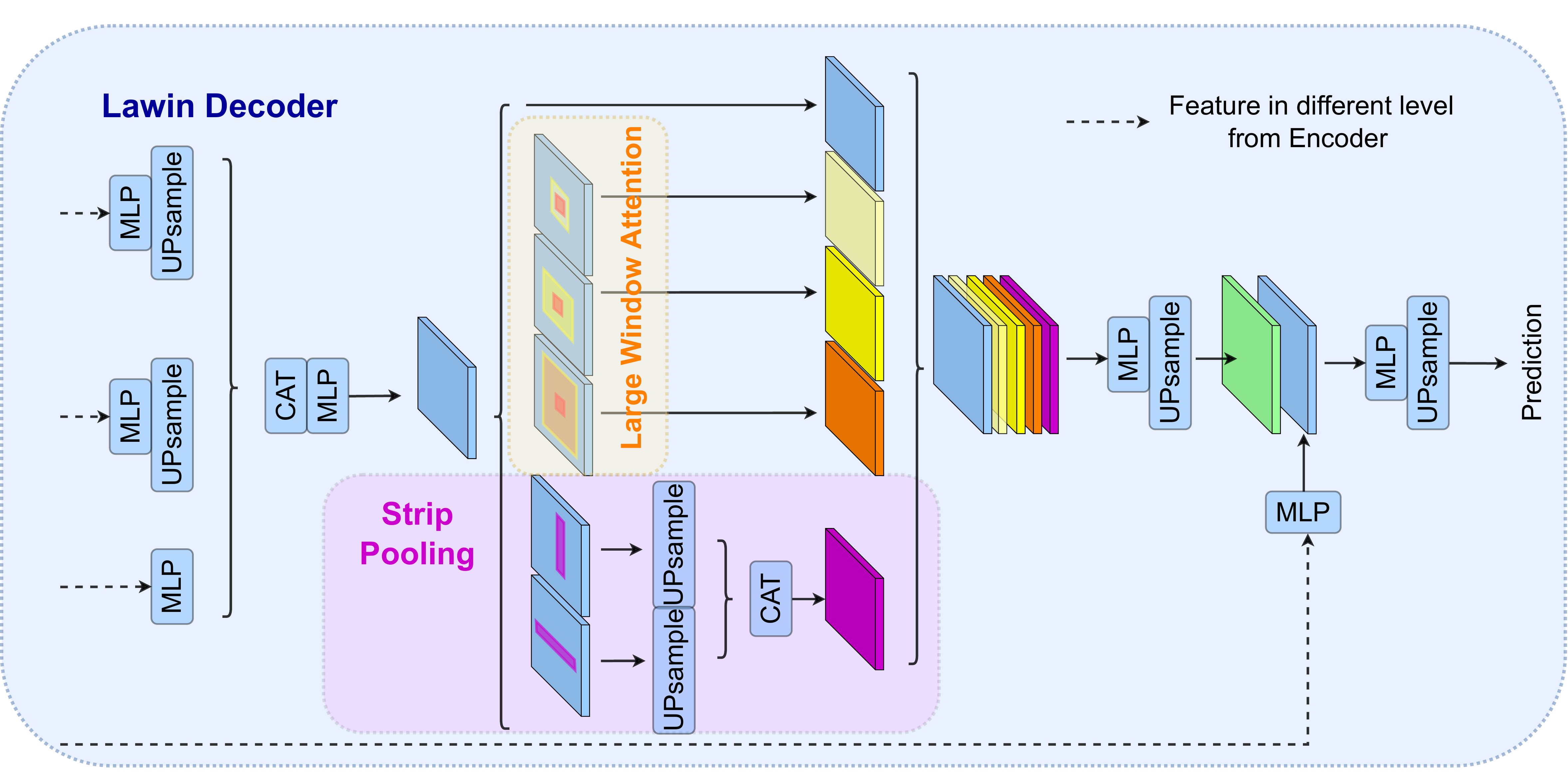}
    \vskip-1ex
    \caption{\textbf{The structure of HybirdASPP} consists of large window attention and strip pooling to capture long-range context.}
    \label{fig:Decoder}
    \vskip-3ex
\end{figure*}

\subsection{HybridASPP}
\label{sec3:3}

Next, we propose multiple improvements to the original SegFormer architecture in order to efficiently extract large regions of the global context and long-range dependencies.
To this intent, our enhanced HybridASPP decoder replaces the vanilla MLP-based decoder of SegFormer.
As shown in Fig.~\ref{fig:Decoder},
in the first yellow branch, HybridASPP inherits the large window attention blocks of Large Window Attention Spatial Pyramid Pooling (LawinASPP)~\cite{yan2022lawin} to exploit global context information.
By adjusting the ratio of the context regions to the query regions with $\{2, 4, 8\}$ as illustrated in Fig.~\ref{fig:Decoder}, large-window attention is able to capture contextual information at multiple scales.
The offered receptive fields are of sizes $\{16, 32, 64\}$, when setting the patch size of the local window to $8$. 
Further, the study in \cite{hou2020strip} found that the standard spatial pooling operation with a shape of $H{\times}W$ makes the long-range context likely to contain many unrelated regions.
In the bottom branch, we intertwine strip pooling modules, which encode long-range context along the horizontal and vertical spatial dimension based on long, narrow kernels ($1{\times}W$ and $H{\times}1$).

As shown in Fig.~\ref{fig:Decoder}, we concatenate an aggregated feature from the last three stages of the encoder, the large window attended features,  the strip pooling augmented feature, after which a learned linear transformation performs dimensionality reduction for producing the final segmentation map.
The output of HybridASPP feature is upsampled to the size of a quarter of input image, then it is fused with the low-level feature coming from the first stage of the segmentation transformer (\ie, SegFormer) via a linear layer.
Last, dense semantic predictions, \ie, the segmentation logits, are obtained from the final representation.

\section{Experiments}
\label{sec:experiments}

\subsection{Datasets}
\label{sec:datasets}
The statistics of the five source datasets  (normal scenes) and the target  dataset (abnormal accident scenes) that are used in our experiments are listed in Table~\ref{tab:stat_datasets}.

\noindent \textbf{Source datasets}. 
For training, we leverage five semantic segmentation datasets as our multi-origin source domains: WildDash2 (\textbf{W}) \cite{wilddash}, ACDC (\textbf{A}) \cite{sakaridis2021acdc}, BDD10K (\textbf{B}) \cite{yu2018bdd100k}, IDD (\textbf{I}) \cite{varma2019idd}, and Cityscapes (\textbf{C}) \cite{cordts2016cityscapes}.
The datasets \textbf{A} and \textbf{B} and \textbf{C} are annotated with the same label set of $19$ categories.
Compared to these datasets, \textbf{W} has $5$ five additional classes, including \emph{van}, \emph{pickup}, \emph{street light}, \emph{billboard}, and \emph{ego-vehicle}.
Following the setting of~\cite{zhang2021generalizable}, we merge the additional categories of \textbf{W} by mapping the added new Ids to the original Ids of \textbf{A}, \textbf{B}, and \textbf{C}.  
Although \textbf{I} also contains more classes, we use the provided public code\footnote{\url{https://github.com/AutoNUE/public-code}} to directly generate the masks with the same label Ids as \textbf{C}.

\begin{table}[!t]
\resizebox{\columnwidth}{!}{
\begin{tabular}{@{}l@{}|c|c|c|c|c|c@{}}
\toprule
\textbf{Datasets} & \textbf{WildDash2} & \textbf{ACDC} & \textbf{BDD} & \textbf{IDD}& \textbf{Cityscapes}  & \textbf{DADA-seg} \\ \midrule
\#training & 2,979 & 1,600 & 7,000 & 6,993 & 2,975 & 12,207\\
\#evaluation & 1,277 & 406 & 1,000 & 981 & 500 & 313\\ 
\bottomrule
\end{tabular}
}
\vskip-1ex
\caption{\small Statistics of datasets for experiments.}
\label{tab:stat_datasets}
\vskip-3ex
\end{table}

\textbf{W} has a large collection of road scenes from different countries, weather and lighting conditions, including $2,979$ training-  and $1,277$ validation images of size $1,920{\times}1,080$.
\textbf{A} includes four common adverse conditions, \ie, fog, nighttime, rain, and snow.
Each of these conditions has $1,000$ images: $400$ for training, $100$ for validation and $500$ for testing (resolution $1,920{\times}1,080$).
\textbf{B} has geographic, environmental, and weather diversity and provides $7,000$ training images and $1,000$ validation images, with a resolution of $1,280{\times}720$.
\textbf{I} features a higher diversity of within-class appearance compared to \textbf{C}.
A total of $10,004$ images with $6,993$ training- and $981$ validation examples are captured from Indian roads with a resolution of $1,280{\times}964$.
The \textbf{C} dataset contains street scenes of $50$ different cities and $19$ categories with high resolution of $2,048{\times}1,080$. The $5,000$ images are divided into $2,975$ training-, $500$ validation-, and $1,525$ test samples.

\noindent \textbf{Target dataset}. 
We utilize  DADA-seg (\textbf{D})~\cite{zhang2020issafe} as our target dataset used for testing our approach, which covers $313$ evaluation images.
Following~\cite{zhang2020issafe,zhang2021exploring}, $12,207$ \textit{unlabelled} images are used for unsupervised  adaptation.
All images are captured in abnormal driving scenes, \ie, \textit{traffic accident scenes}. The images of \textbf{D} are labelled with $19$ classes, which are consistent with the classes of the \textbf{C} dataset (Cityscapes). The resolution of the images is $1,584{\times}660$.

\begin{table*}[t]
\begin{center}
\resizebox{\textwidth}{!}{
\begin{tabular}{@{}l@{}|l@{}|c|ccccccccccccccccccc@{}}
\toprule
&Method & mIoU & \rotatebox{90}{road} & \rotatebox{90}{sidewalk} & \rotatebox{90}{building} & \rotatebox{90}{wall} & \rotatebox{90}{fence} & \rotatebox{90}{pole} & \rotatebox{90}{traffic light} & \rotatebox{90}{traffic sign} & \rotatebox{90}{vegetation} & \rotatebox{90}{terrain} & \rotatebox{90}{sky} & \rotatebox{90}{person} & \rotatebox{90}{rider} & \rotatebox{90}{car} & \rotatebox{90}{truck} & \rotatebox{90}{bus} & \rotatebox{90}{train} & \rotatebox{90}{motorcycle} &
\rotatebox{90}{bicycle}\\
\midrule\midrule
{\multirow{10}{*}{\rotatebox[origin=c]{90}{\textit{Source-only}}}} 
&MobileNetV2~\cite{sandler2018mobilenetv2} &16.05& 31.87&8.50&26.55&3.60&5.38&13.96&19.51&10.87&44.99&11.09&67.05&8.11&5.23&28.58&11.77&2.17&-&1.90&3.86\\
&PSPNet~\cite{Zhao_2017_CVPR} & 17.07&31.62 & 11.42 & 32.48 & 4.16 & 8.52 & 12.38 & 17.93 & 13.39 & 50.82 & 13.85 & 67.19 & 9.86 & 3.13 & 31.54 & 6.97 & 3.15 & - & 2.97 & 2.89\\
&ResNet50~\cite{he2016resnet} &18.96&34.19 & 8.24 & 31.05 & 4.56 & 7.39 & 19.04 & 27.05 & 15.35 & 33.30 & 12.40 & 61.52 & 10.04 & 3.95 & 42.59 & 14.15 & 27.02 & - & 3.72 & 4.72\\
&SemFPN~\cite{kirillov2019semanticfpn}&19.59&37.90 & 10.12 & 23.80 & 3.74 & 9.64 & 22.06 & 28.64 & 15.55 & 40.95 & 12.13 & 51.93 & 9.24 & 5.93 & 52.08 & 13.89 & 26.54 & - & 3.66 & 4.36\\
&DNLNet~\cite{yin2020dnl}& 19.72&41.68 & 13.26 & 30.45 & 6.17 & 11.04 & 21.91 & 28.03 & 17.99 & 40.05 & 14.13 & 56.06 & 10.75 & 5.41 & 34.78 & 8.01 & 28.01 & - & 3.55 & 3.39\\
&ResNeSt~\cite{zhang2020resnest}&19.99&39.63 & 11.38 & 33.68 & 2.81 & 9.73 & 22.76 & 27.35 & 18.09 & 45.24 & 14.22 & 71.23 & 13.34 & 5.03 & 36.45 & 6.91 & 13.08 & - & 3.94 & 4.87\\
&DANet~\cite{fu2019dual}&22.24&46.49 & 10.17 & 42.20 & 3.81 & 10.65 & 13.46 & 18.69 & 22.59 & 55.76 & 22.22 & 83.84 & 6.68 & 11.75 & 39.59 & 7.96 & 12.64 & - & 7.98 & 6.12\\
&ResNet101~\cite{he2016resnet} &23.60&57.96 & 11.16 & 39.94 & 6.43 & 9.46 & 23.67 & 27.37 & 17.32 & 45.65 & 16.47 & 69.21 & 13.19 & 4.51 & 47.29 & 13.75 & 30.44 & - & 6.64 & 8.01\\
&OCRNet~\cite{yuan2020object_contextual_representations}&24.85&42.13 & 11.54 & 34.49 & 6.63 & 12.70 & 22.76 & 29.03 & 22.28 & 42.41 & 15.15 & 85.43 & 14.31 & 6.65 & 53.94 & 20.65 & 34.86 & - & 9.30 & 7.87\\
&FastSCNN~\cite{poudel2019fast}&26.32&69.91 & 16.30 & 52.53 & 6.09 & 9.63 & 19.98 & 19.30 & 22.58 & 57.04 & 22.95 & 90.81 & 11.19 & 13.95 & 46.16 & 22.65 & 9.74 & - & 4.49 & 4.75\\ \midrule

{\multirow{9}{*}{\rotatebox[origin=c]{90}{\textit{Cross-source}}}} 
&CLAN~\cite{luo2019CLAN} & 28.76 & 79.80 & 18.61 & 51.56 & 8.32 & 13.60 & 15.51 & 17.15 & 21.51 & 63.20 & 21.99 & 80.53 & 8.37 & 6.32 & 63.47 & 33.43 & 33.12 & - & 3.69 & 6.21\\
&BDL~\cite{li2019bidirectional} & 29.66 & 81.44 & 19.18 & 57.18 & 8.61 & 16.26 & 14.65 & 8.78 & 16.77 & 66.60 & 26.83 & 85.87 & 10.51 & 7.16 & 65.45 & 35.18 & 34.78 & - & 2.71 & 5.57\\
&ISSAFE~\cite{zhang2020issafe} & 29.97 & 80.23 & 19.51 & 52.02 & 6.43 & 14.68 & 16.19 & 17.03 & 19.50 & 65.39 & 21.69 & 79.84 & 9.95 & 8.82 & 65.60 & 39.51 & 39.73 & - & 6.09 & 7.03\\
&EDCNet~\cite{zhang2021exploring} &32.04 & 73.03 & 19.47 & 57.31 & 11.60 & 14.30 & 20.70 & 12.27 & 27.22 & 70.54 & 18.98 & 88.64 & 10.69 & 8.70 & 68.14 & 49.80 & 50.86 & - & 9.02 & 4.12\\
&Trans4Trans-M~\cite{zhang2021trans4trans} & 39.20 & 71.10 &15.57 &70.39 &10.34 &16.53 &31.63 & \textbf{37.16} & \textbf{37.38} &71.88 &19.61 &93.04 &21.27 &14.97 &64.04 &53.76 &81.53 &- &24.63 &10.07 \\
\cline{2-22}
&Our Baseline & 40.73 & 84.93 & 23.66 & 68.34 & 16.27 & 20.58 & 25.96 & 31.25 & 28.20 & 71.89 & 22.39 & 93.16 & 17.92 & 26.84 & 73.89 & 55.09 & 69.26 & - & 34.77 & 9.49\\ 
& \quad+Meta & 45.03 & 86.20 & 25.44 & 70.63 & 14.21 & 19.75 & 26.56 & 28.01 & 29.23 & 74.45 & 25.29 & 93.18 & 20.40 & 31.53 & 75.02 & 64.73 & 76.84 & - & 38.05 & 10.95
\\
&  \quad+Meta+MDMS & 46.11 & 87.10 & 27.71 & 71.11 & \textbf{22.94} & \textbf{20.64} & \textbf{32.25} & 29.49 & 34.34 & 75.48 & 24.02 & 92.18 & 20.65 & \textbf{33.33} & 74.64 & 63.35 & 71.14 & - & \textbf{39.08} & \textbf{12.04}
\\ 
&Our MMUDA  & \textbf{46.97} & \textbf{87.51} & \textbf{27.97} & \textbf{74.76} & 16.16 & 21.93 & 29.94 & 29.43 & 31.62 & \textbf{75.67} & \textbf{26.69} & \textbf{93.57} & \textbf{24.40} & 29.57 & \textbf{77.35} & \textbf{68.24} & \textbf{84.02} & - & 36.96 & 10.44
\\
\bottomrule
\end{tabular}}
\end{center}
\vskip-3ex
\caption{\textbf{Comparison of state-of-art methods.} The source-only models are trained on the Cityscapes dataset, while the other models are domain-transferred using a single source~\cite{luo2019CLAN,li2019bidirectional}, multiple sources~\cite{zhang2021exploring,zhang2021trans4trans} or a different modality~\cite{zhang2020issafe}. While our baseline model has no mixed sampling and performs normal supervised learning based on ResNet101 only using the source domain, our MMUDA framework based on the SegFormer backbone uses MDMS and meta-learning (Meta for short) across the source and target domain.}
\label{tab:sota_densepass}
\vskip-4ex
\end{table*}

\subsection{Implementation Details}
Our approach leverages the ImageNet1K-pretrained MiT-B2 SegFormer~\cite{xie2021segformer} as the encoder backbone, and the public mmsegmentation framework implementation\footnote{\url{https://github.com/open-mmlab/mmsegmentation}}.
The meta-learning inner and outer learning rates (\ie $\eta$ and $\gamma$)  are set to $1e^{-3}$ and $5e^{-3}$, respectively, with Polynomial learning rate decay with the power $0.9$.
The network is trained for $120$ epochs, unless otherwise specified.
The weight $\alpha$ of the $\mathcal{L}_{ds}$ loss is set to $1$. 
We train the model with a mini-batch size of $1$ using stochastic gradient descent (SGD),  momentum of $0.9$ and weight decay of $5e^{-4}$. 
For training data augmentation, we use random resizing with ratio $0.5$ to $2.0$, random flipping, random Gaussian blur, and random cropping with a size of $600{\times}600$. Mean Intersection over Union (mIoU) is our main evaluation metric.

\subsection{Results of Accident Scenes Segmentation}
Table~\ref{tab:sota_densepass} compares the mean IoU and the per-class IoU scores achieved by our proposed approach to  the state-of-the-art methods on DADA-seg.
The source-only models trained on Cityscapes experience a considerable   performance degradation and achieve a rather low accuracy when deployed in abnormal accident scenes.
For example, the ResNet101~\cite{he2016resnet}, whose results are illustrated in Fig.~\ref{fig:1}, only reaches $23.60\%$ in mIoU.
The previous state-of-the-art Trans4Trans model~\cite{zhang2021trans4trans} attains $39.20\%$ in mIoU with a vision transformer and multi-source training.
Our proposed MMUDA model surpasses all previous results, yielding a significantly  higher recognition rate of $46.97\%$ in mIoU, which is  ${>}7.50\%$ higher than the past state-of-the-art.
For per-class IoU, our approach achieves the best scores in $16$ out of all $19$ categories. 
The improvements over Trans4Trans are compelling (${>}10.00\%$ performance gain) on categories which are safety-critical for accidental scene understanding, in particular, for \emph{road}, \emph{sidewalk}, \emph{rider}, \emph{car}, \emph{truck}, and \emph{motorcycle}. 

\begin{figure*}
    \centering
    \includegraphics[width=\textwidth]{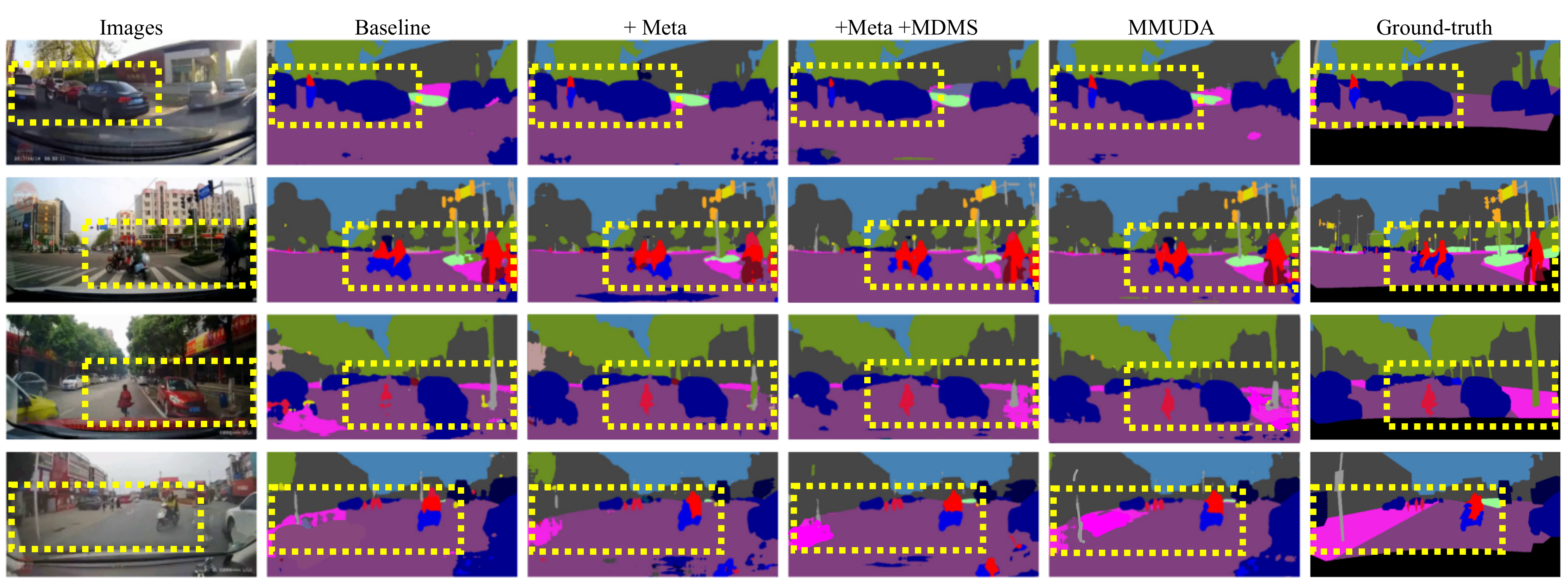}
    \vskip-2ex
    \caption{\textbf{Qualitative ablation study} of different modules used in our framework.}
    \label{fig:vis_ablation}
    \vskip-3ex
\end{figure*}

The ablation results of our proposed modules are also depicted in Table~\ref{tab:sota_densepass}, where all experiments are based on all five multi-origin source datasets described in Sec.~\ref{sec:datasets}.
Our baseline model with ResNet101 uses only the normal source-supervised learning by aggregating multiple source domains and obtains a mIoU of $40.76\%$, whereas the model with meta-learning (+Meta) leads to a further $4.26\%$ improvement.
In addition, our proposed MDMS and transformer model with HyrbidASPP further elevate mIoU to $46.11\%$ and $46.97\%$, respectively.
Overall, these results demonstrate the effectiveness of the proposed framework and clear benefits of the MDMS module and the transformer model with HybridASPP, as well as the importance of our multi-source meta-learning training strategy.

\begin{table}[t]
\centering
\resizebox{.8\columnwidth}{!}{
    \begin{tabular}{c|c|c}
        \toprule
        \multirow{2}*{\textbf{Source domain}} &  \multicolumn{2}{c}{\textbf{\acrshort{miou}}(\%)} \\ \cline{2-3}
         & Meta & without Meta \\ \hline\hline
        \textbf{W}ildDash2~\cite{wilddash} & - & 35.35 \\ 
        \textbf{A}CDC~\cite{sakaridis2021acdc} & - & 23.56 \\
        \textbf{B}DD~\cite{yu2018bdd100k}& - & 31.18\\
        \textbf{I}DD~\cite{varma2019idd} & - & 28.15\\
        \textbf{C}ityscapes~\cite{cordts2016cityscapes} & - & 25.48\\ \hline
        \textbf{W} + \textbf{I} + \textbf{C} & 43.74  & 40.01  \\ 
        \textbf{W} + \textbf{A} + \textbf{B} & 42.70 & 38.75 \\ 
        \textbf{W} + \textbf{A} + \textbf{I} & 43.36 & 39.98 \\ 
        \textbf{W} + \textbf{B} + \textbf{I} & 43.90 & 40.20 \\ 
        \textbf{B} + \textbf{I} + \textbf{C} & 41.03 & 39.03 \\ 
        \textbf{W} + \textbf{B} + \textbf{I} + \textbf{C} & 44.67 & 40.71 \\ \hline
        \textbf{W} + \textbf{A} + \textbf{B} + \textbf{I} + \textbf{C} & 45.03 & 40.73 \\ 
        \bottomrule
    \end{tabular}}
\vskip-1ex
\caption{\small \textbf{Ablation of meta-learning} with different source domains and their combinations.}
\label{tab:meta-learning}
\vskip-4ex
\end{table}

\subsection{Ablation Study of Meta-learning}

\noindent \textbf{Effect of meta-learning.} The previously described Table~\ref{tab:sota_densepass} indicates clear advantages of meta-learning over the baseline.
Next, we study the impact of our meta-learning strategy in more detail though an ablation study featuring a variety of source datasets (Table~\ref{tab:meta-learning}).
All of these experiments employ the target-specific normalization and ResNet101 pretained on ImageNet1K~\cite{imagenet} as the backbone.
The baseline models are trained with traditional learning on different single sources, respectively. 
When using multiple sources, meta-learning-based approaches (\eg, $43.74\%$ with \textbf{W}+\textbf{I}+\textbf{C}) are evidently more effective than the common aggregated-source-supervised learning (\eg, $40.01\%$).

\noindent \textbf{Influence of multi-source combination.}
In Table~\ref{tab:meta-learning}, we investigate the impact of combining different datasets on the segmentation performance.
In single-source studies, training with \textbf{W}ildDash2, \textbf{B}DD, and \textbf{I}DD, respectively, yields the top three \acrshort{miou} scores thanks to their especially diverse examples.
For example, WildDash2 leads to a decent performance, as it offers many composite scenes and various visual hazards.
As a result, when these three sources are used together, the combination of \textbf{W}+\textbf{B}+\textbf{I} yields the highest accuracy.
In general, leaning with more sources improves the  performance for accident scene segmentation, and our five-source meta-learning model attains $45.03\%$ in mIoU, clearly standing out in front of all other  models.

\begin{table}[t]
\centering
\resizebox{.9\columnwidth}{!}{
    \begin{tabular}{l|c|c|c}
        \toprule
        \multirow{2}*{\textbf{Method}} & \multirow{2}*{\textbf{GFLOPs}$\downarrow$} & \multicolumn{2}{c}{\textbf{\acrshort{miou}}(\%)$\uparrow$} \\ \cline{3-4}
        \quad & & \textbf{C}ityscapes & \textbf{D}ADA-seg \\ \hline\hline
        SegFormer + Vanilla MLP~\cite{xie2021segformer}& 717.1 & 74.00 & 18.50\\  
        SegFormer + LawinASPP~\cite{yan2022lawin} & 569.6 & 75.86 & 25.16\\
        SegFormer + HybridASPP (Ours) & 553.8 & 76.41 & 25.21\\
        \bottomrule
    \end{tabular}}
    \vskip-1ex
\caption{\small \textbf{Ablation of decoders.} GFLOPs are calculated with a size of $2,048{\times}1,024$. We train all models on \textbf{C} on a single GPU with a batch size of 1 for $80k$ iterations and an input size of $768{\times}768$.}
\label{tab:abl_hybridaspp}
\vskip-3ex
\end{table}
\subsection{Ablation Study on HybridASPP}
To analyze the efficiency and effectiveness of our HybridASPP module, we replace the decoder of SegFormer-B2~\cite{xie2021segformer}, and compare it to LawinASPP~\cite{yan2022lawin} and the proposed decoder in Table~\ref{tab:abl_hybridaspp}.
In terms of computation costs, HybridASPP reduces the GFLOPs of LawinASPP by $15.8$ after introducing strip pooling to capture long-range context instead of using traditional image pooling.
Looking at the performance, our decoder on \textbf{D} is evidently more reliable than Vanilla MLP.
Meanwhile, our HybridASPP achieves higher \acrshort{miou} scores than LawinASPP with less GFLOPs.

\subsection{Model Efficiency Analysis}
To investigate the efficiency of MMUDA, we compare it with state-of-the-art approaches evaluated on DADA-seg following~\cite{zhang2021trans4trans}.
Our model employs MiT-B2~\cite{xie2021segformer} as the encoder and the proposed HybridASPP as the decoder.
The comparison of \acrshort{miou}, GFLOPs, and \#Params is shown in Table~\ref{tab:efficiency}.
Compared to DeepLabV3+~\cite{chen2018deeplabv3+} and HRNet~\cite{hrnet}, our model largely improves the performance, while saving a great amount of computation demands. In comparison with the previous state-of-the-art Trans4Trans-M, we obtain a significant gain of $7.8\%$ in \acrshort{miou} with $19.1M$ less parameters and an increase in GFLOPs.
Capturing rich contextual cues requires more computation but also ensures the robustness of the model. 
While our model achieves high efficiency in general, in future work, we will consider a more light-weight encoder to further reduce the computational cost.

\begin{table}[t]
\centering
\resizebox{.9\columnwidth}{!}{%
    \begin{tabular}{l|ccc}
        \toprule
        \textbf{Method} & \textbf{\acrshort{miou}} (\%) $\uparrow$ & \textbf{GFLOPs} $\downarrow$ & \textbf{\#Params (M)} $\downarrow$ \\ \hline\hline
        DeepLabV3+~\cite{chen2018deeplabv3+} & 26.8 & 178.1 & 18.70\\
        HRNet~\cite{hrnet} & 27.5 & 210.5 & 65.86 \\
        Trans4Trans-M~\cite{zhang2021trans4trans} & 39.2 & 41.9 & 49.55\\ 
        \hline
        Ours & 47.0 & 105.5 & 30.49\\
        \bottomrule
    \end{tabular}
}
\vskip-1ex
\caption{\small \textbf{Model efficiency analysis.} GFLOPs are calculated with the input size of $768{\times}768$.}
\label{tab:efficiency}
\vskip-3ex
\end{table}
        
\begin{figure*}
    \centering
    \includegraphics[width=\textwidth]{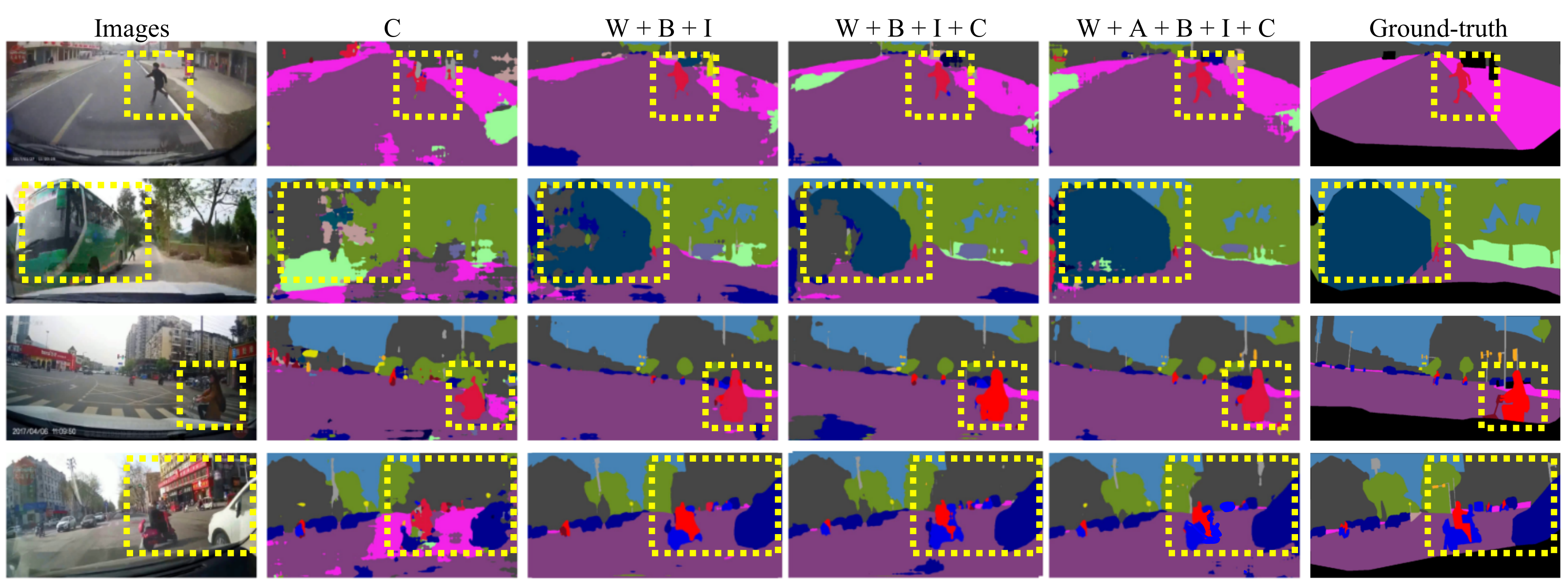}
    \vskip-2ex
    \caption{\textbf{Qualitative analysis} of the effect of using different numbers of source domains in our multi-source meta-learning framework.}
    \label{fig:vis_multi_source}
    \vskip-3ex
\end{figure*}

\subsection{Qualitative Analysis}
\noindent \textbf{Effect of MMUDA.}
In our final study, we showcase multiple examples of representative qualitative results in  Fig.~\ref{fig:vis_ablation},  which corresponds to the numerical results in Table~\ref{tab:sota_densepass}.
Our baseline model produces relatively noisy segmentation results, such that the rushing person in the third row can not be completely segmented, whereas the model using meta-learning method makes more accurate predictions in this regard.
When a collision happens, the model with the additive MDMS module is better at distinguishing between the \emph{motorcyclist} and the \emph{motorcycle}.
Furthermore, the segmentation of \textit{sidewalk} is also improved.
In comparison, the noise of the predictions made by our MMUDA model with HyridASPP is much lower and the model segments the sidewalks well even under low light and with occlusion.
Overall, our model adapts well to the accident scenario and is capable of providing robustness-improved semantic segmentation for the safety of autonomous driving.

\noindent \textbf{Effect of source data.} Fig.~\ref{fig:vis_multi_source} visualizes the performance of our model when using different numbers of source domains. The images contain distortion and blur of foreground objects.
It can be seen that the Cityscapes-trained model yields fragmented segmentation that disqualifies its application in self-driving scenarios, as it poses great threats in potential accident scenes. 
Comparing the predictions in the dashed box, our MMUDA provides more robust segmentation results when intertwining more sources.
Since DADA-seg is a complex driving scenario, the model will be more effective when the sources provide more diverse information.
The five-source meta-learned model clearly robustifies and improves semantic segmentation of accident scenes, and we believe that the high-quality predictions can be propagated to the upper-level driving applications.

\section{Conclusion}
\label{sec:conclusion}

Semantic segmentation of road scenes is a key ingredient for safe autonomous driving, but requires models that reliably operate under unusual circumstances.
In this work, we specifically focus on segmentation of \textit{abnormal} accident scenes since unexpected objects or traffic scenarios forms one of the common cause of dangerous situations.
To tackle this challenge, we introduce a new framework which transfers knowledge from the well-studied domain of \textit{standard} image segmentation to our target domain of \textit{abnormal} scenes.
Our \emph{Multi-source Meta-learning UDA (MMUDA)} framework leverages multi-domain mixed sampling targeting at a better adaptation to the unknown accident scenes and is optimized using meta-learning.
We further introduce a HybridASPP decoder design to improve the SegFormer segmentation backbone, which proved to be effective for our task.
We verify the robustness of our model through extensive quantitative and qualitative experiments on the public DADA-seg benchmark, demonstrating superior generalization ability to abnormal accident scenes and surpassing previous state-of-the-art by a large margin.

\noindent \textbf{Limitation and broader impact.} The model training is limited by not using the augmented set as the meta-train set. To address this issue, a potential approach is to combine multiple source domains and target domain into a fusion set, and then to conduct a cross-combination meta-learning process. We leave it to our further research.
Besides, the current experiments are conducted based on the referred datasets, thus there are still data biases when the model is applied in real-world test fields.

\clearpage
{\small
\bibliographystyle{ieee_fullname}
\bibliography{egbib}

\begin{thebibliography}{10}\itemsep=-1pt

\bibitem{balaji2018metareg}
Yogesh Balaji, Swami Sankaranarayanan, and Rama Chellappa.
\newblock {MetaReg:} {Towards} domain generalization using meta-regularization.
\newblock In {\em NeurIPS}, 2018.

\bibitem{chen2017deeplab}
Liang{-}Chieh Chen, George Papandreou, Iasonas Kokkinos, Kevin Murphy, and
  Alan~L. Yuille.
\newblock {DeepLab:} {Semantic} image segmentation with deep convolutional
  nets, atrous convolution, and fully connected {CRFs}.
\newblock {\em TPAMI}, 2018.

\bibitem{chen2018deeplabv3+}
Liang-Chieh Chen, Yukun Zhu, George Papandreou, Florian Schroff, and Hartwig
  Adam.
\newblock Encoder-decoder with atrous separable convolution for semantic image
  segmentation.
\newblock In {\em ECCV}, 2018.

\bibitem{chen2021cyclemlp}
Shoufa Chen, Enze Xie, Chongjian Ge, Ding Liang, and Ping Luo.
\newblock {CycleMLP:} {A} {MLP-like} architecture for dense prediction.
\newblock In {\em ICLR}, 2022.

\bibitem{chen2019blending}
Ziliang Chen, Jingyu Zhuang, Xiaodan Liang, and Liang Lin.
\newblock Blending-target domain adaptation by adversarial meta-adaptation
  networks.
\newblock In {\em CVPR}, 2019.

\bibitem{cheng2021maskformer}
Bowen Cheng, Alex Schwing, and Alexander Kirillov.
\newblock Per-pixel classification is not all you need for semantic
  segmentation.
\newblock In {\em NeurIPS}, 2021.

\bibitem{Choi_2021_CVPR}
Sungha Choi, Sanghun Jung, Huiwon Yun, Joanne~T. Kim, Seungryong Kim, and
  Jaegul Choo.
\newblock {RobustNet:} {Improving} domain generalization in urban-scene
  segmentation via instance selective whitening.
\newblock In {\em CVPR}, 2021.

\bibitem{cordts2016cityscapes}
Marius Cordts, Mohamed Omran, Sebastian Ramos, Timo Rehfeld, Markus Enzweiler,
  Rodrigo Benenson, Uwe Franke, Stefan Roth, and Bernt Schiele.
\newblock The cityscapes dataset for semantic urban scene understanding.
\newblock In {\em CVPR}, 2016.

\bibitem{imagenet}
Jia Deng, Wei Dong, Richard Socher, Li-Jia Li, Kai Li, and Li Fei-Fei.
\newblock {ImageNet:} {A} large-scale hierarchical image database.
\newblock In {\em CVPR}, 2009.

\bibitem{dou2019domain_generalization}
Qi Dou, Daniel Coelho~de Castro, Konstantinos Kamnitsas, and Ben Glocker.
\newblock Domain generalization via model-agnostic learning of semantic
  features.
\newblock In {\em NeurIPS}, 2019.

\bibitem{finn2017model}
Chelsea Finn, Pieter Abbeel, and Sergey Levine.
\newblock Model-agnostic meta-learning for fast adaptation of deep networks.
\newblock In {\em ICML}, 2017.

\bibitem{fu2019dual}
Jun Fu, Jing Liu, Haijie Tian, Yong Li, Yongjun Bao, Zhiwei Fang, and Hanqing
  Lu.
\newblock Dual attention network for scene segmentation.
\newblock In {\em CVPR}, 2019.

\bibitem{gong2021cluster}
Rui Gong, Yuhua Chen, Danda~Pani Paudel, Yawei Li, Ajad Chhatkuli, Wen Li,
  Dengxin Dai, and Luc Van~Gool.
\newblock Cluster, split, fuse, and update: Meta-learning for open compound
  domain adaptive semantic segmentation.
\newblock In {\em CVPR}, 2021.

\bibitem{ha2017hypernetworks}
David Ha, Andrew~M. Dai, and Quoc~V. Le.
\newblock Hypernetworks.
\newblock In {\em ICLR}, 2017.

\bibitem{He_2021_CVPR}
Jianzhong He, Xu Jia, Shuaijun Chen, and Jianzhuang Liu.
\newblock Multi-source domain adaptation with collaborative learning for
  semantic segmentation.
\newblock In {\em CVPR}, 2021.

\bibitem{he2016resnet}
Kaiming He, Xiangyu Zhang, Shaoqing Ren, and Jian Sun.
\newblock Deep residual learning for image recognition.
\newblock In {\em CVPR}, 2016.

\bibitem{hou2022vision_permutator}
Qibin Hou, Zihang Jiang, Li Yuan, Ming-Ming Cheng, Shuicheng Yan, and Jiashi
  Feng.
\newblock Vision permutator: A permutable {MLP-like} architecture for visual
  recognition.
\newblock {\em TPAMI}, 2022.

\bibitem{hou2020strip}
Qibin Hou, Li Zhang, Ming-Ming Cheng, and Jiashi Feng.
\newblock Strip pooling: Rethinking spatial pooling for scene parsing.
\newblock In {\em CVPR}, 2020.

\bibitem{Huang_2021_CVPR}
Jiaxing Huang, Dayan Guan, Aoran Xiao, and Shijian Lu.
\newblock {FSDR:} {Frequency} space domain randomization for domain
  generalization.
\newblock In {\em CVPR}, 2021.

\bibitem{kirillov2019semanticfpn}
Alexander Kirillov, Ross Girshick, Kaiming He, and Piotr Doll{\'a}r.
\newblock Panoptic feature pyramid networks.
\newblock In {\em CVPR}, 2019.

\bibitem{li2020online}
Da Li and Timothy Hospedales.
\newblock Online meta-learning for multi-source and semi-supervised domain
  adaptation.
\newblock In {\em ECCV}, 2020.

\bibitem{li2018learning}
Da Li, Yongxin Yang, Yi{-}Zhe Song, and Timothy~M. Hospedales.
\newblock Learning to generalize: Meta-learning for domain generalization.
\newblock In {\em AAAI}, 2018.

\bibitem{li2019bidirectional}
Yunsheng Li, Lu Yuan, and Nuno Vasconcelos.
\newblock Bidirectional learning for domain adaptation of semantic
  segmentation.
\newblock In {\em CVPR}, 2019.

\bibitem{liu2020open}
Ziwei Liu, Zhongqi Miao, Xingang Pan, Xiaohang Zhan, Dahua Lin, Stella~X. Yu,
  and Boqing Gong.
\newblock Open compound domain adaptation.
\newblock In {\em CVPR}, 2020.

\bibitem{long2015fully}
Jonathan Long, Evan Shelhamer, and Trevor Darrell.
\newblock Fully convolutional networks for semantic segmentation.
\newblock In {\em CVPR}, 2015.

\bibitem{luo2019CLAN}
Yawei Luo, Liang Zheng, Tao Guan, Junqing Yu, and Yi Yang.
\newblock Taking a closer look at domain shift: Category-level adversaries for
  semantics consistent domain adaptation.
\newblock In {\em CVPR}, 2019.

\bibitem{olsson2021classmix}
Viktor Olsson, Wilhelm Tranheden, Juliano Pinto, and Lennart Svensson.
\newblock {ClassMix:} {Segmentation-based} data augmentation for
  semi-supervised learning.
\newblock In {\em WACV}, 2021.

\bibitem{orsic2019defense}
Marin Orsic, Ivan Kreso, Petra Bevandic, and Sinisa Segvic.
\newblock In defense of pre-trained {ImageNet} architectures for real-time
  semantic segmentation of road-driving images.
\newblock In {\em CVPR}, 2019.

\bibitem{pan2018two_at_once}
Xingang Pan, Ping Luo, Jianping Shi, and Xiaoou Tang.
\newblock Two at once: Enhancing learning and generalization capacities via
  {IBN-net}.
\newblock In {\em ECCV}, 2018.

\bibitem{park2020discover}
Kwanyong Park, Sanghyun Woo, Inkyu Shin, and In~So Kweon.
\newblock Discover, hallucinate, and adapt: Open compound domain adaptation for
  semantic segmentation.
\newblock In {\em NeurIPS}, 2020.

\bibitem{peng2021mass}
Kunyu Peng, Juncong Fei, Kailun Yang, Alina Roitberg, Jiaming Zhang, Frank
  Bieder, Philipp Heidenreich, Christoph Stiller, and Rainer Stiefelhagen.
\newblock {MASS:} {Multi-attentional} semantic segmentation of {LiDAR} data for
  dense top-view understanding.
\newblock {\em T-ITS}, 2022.

\bibitem{poudel2019fast}
Rudra P.~K. Poudel, Stephan Liwicki, and Roberto Cipolla.
\newblock {Fast-SCNN:} {Fast} semantic segmentation network.
\newblock In {\em BMVC}, 2019.

\bibitem{romera2017erfnet}
Eduardo Romera, Jose~M. Alvarez, Luis~Miguel Bergasa, and Roberto Arroyo.
\newblock {ERFNet:} {Efficient} residual factorized {ConvNet} for real-time
  semantic segmentation.
\newblock {\em T-ITS}, 2018.

\bibitem{romera2019bridging}
Eduardo Romera, Luis~Miguel Bergasa, Kailun Yang, Jose~M. Alvarez, and Rafael
  Barea.
\newblock Bridging the day and night domain gap for semantic segmentation.
\newblock In {\em IV}, 2019.

\bibitem{sakaridis2021acdc}
Christos Sakaridis, Dengxin Dai, and Luc Van~Gool.
\newblock {ACDC:} {The} adverse conditions dataset with correspondences for
  semantic driving scene understanding.
\newblock In {\em ICCV}, 2021.

\bibitem{saleh2018effective}
Fatemeh~Sadat Saleh, Mohammad~Sadegh Aliakbarian, Mathieu Salzmann, Lars
  Petersson, and Jose~M. Alvarez.
\newblock Effective use of synthetic data for urban scene semantic
  segmentation.
\newblock In {\em ECCV}, 2018.

\bibitem{sandler2018mobilenetv2}
Mark Sandler, Andrew Howard, Menglong Zhu, Andrey Zhmoginov, and Liang-Chieh
  Chen.
\newblock {MobilenetV2:} {Inverted} residuals and linear bottlenecks.
\newblock In {\em CVPR}, 2018.

\bibitem{strudel2021segmenter}
Robin Strudel, Ricardo Garcia, Ivan Laptev, and Cordelia Schmid.
\newblock Segmenter: Transformer for semantic segmentation.
\newblock In {\em ICCV}, 2021.

\bibitem{sun2019see}
Lei Sun, Kaiwei Wang, Kailun Yang, and Kaite Xiang.
\newblock See clearer at night: {Towards} robust nighttime semantic
  segmentation through day-night image conversion.
\newblock In {\em SPIE}, 2019.

\bibitem{tolstikhin2021mlp}
Ilya~O. Tolstikhin, Neil Houlsby, Alexander Kolesnikov, Lucas Beyer, Xiaohua
  Zhai, Thomas Unterthiner, Jessica Yung, Andreas Steiner, Daniel Keysers,
  Jakob Uszkoreit, Mario Lucic, and Alexey Dosovitskiy.
\newblock {MLP-mixer:} {An} {all-MLP} architecture for vision.
\newblock In {\em NeurIPS}, 2021.

\bibitem{tranheden2021dacs}
Wilhelm Tranheden, Viktor Olsson, Juliano Pinto, and Lennart Svensson.
\newblock {DACS:} {Domain} adaptation via cross-domain mixed sampling.
\newblock In {\em WACV}, 2021.

\bibitem{varma2019idd}
Girish Varma, Anbumani Subramanian, Anoop~M. Namboodiri, Manmohan Chandraker,
  and C.~V. Jawahar.
\newblock {IDD:} {A} dataset for exploring problems of autonomous navigation in
  unconstrained environments.
\newblock In {\em WACV}, 2019.

\bibitem{hrnet}
Jingdong Wang, Ke Sun, Tianheng Cheng, Borui Jiang, Chaorui Deng, Yang Zhao,
  Dong Liu, Yadong Mu, Mingkui Tan, Xinggang Wang, Wenyu Liu, and Bin Xiao.
\newblock Deep high-resolution representation learning for visual recognition.
\newblock {\em TPAMI}, 2021.

\bibitem{wang2022dual}
Ying Wang, Chiuman Ho, Wenju Xu, Ziwei Xuan, Xudong Liu, and Guo-Jun Qi.
\newblock Dual-flattening transformers through decomposed row and column
  queries for semantic segmentation.
\newblock {\em arXiv preprint arXiv:2201.09139}, 2022.

\bibitem{xie2021segformer}
Enze Xie, Wenhai Wang, Zhiding Yu, Anima Anandkumar, Jose~M. Alvarez, and Ping
  Luo.
\newblock {SegFormer:} {Simple} and efficient design for semantic segmentation
  with transformers.
\newblock In {\em NeurIPS}, 2021.

\bibitem{yan2022lawin}
Haotian Yan, Chuang Zhang, and Ming Wu.
\newblock Lawin transformer: Improving semantic segmentation transformer with
  multi-scale representations via large window attention.
\newblock {\em arXiv preprint arXiv:2201.01615}, 2022.

\bibitem{yang2021bi}
Guanglei Yang, Zhun Zhong, Hao Tang, Mingli Ding, Nicu Sebe, and Elisa Ricci.
\newblock {Bi-mix:} {Bidirectional} mixing for domain adaptive nighttime
  semantic segmentation.
\newblock {\em arXiv preprint arXiv:2111.10339}, 2021.

\bibitem{yang2021capturing}
Kailun Yang, Jiaming Zhang, Simon Rei{\ss}, Xinxin Hu, and Rainer Stiefelhagen.
\newblock Capturing omni-range context for omnidirectional segmentation.
\newblock In {\em CVPR}, 2021.

\bibitem{yin2020dnl}
Minghao Yin, Zhuliang Yao, Yue Cao, Xiu Li, Zheng Zhang, Stephen Lin, and Han
  Hu.
\newblock Disentangled non-local neural networks.
\newblock In {\em ECCV}, 2020.

\bibitem{yu2018bdd100k}
Fisher Yu, Haofeng Chen, Xin Wang, Wenqi Xian, Yingying Chen, Fangchen Liu,
  Vashisht Madhavan, and Trevor Darrell.
\newblock {BDD100K:} {A} diverse driving dataset for heterogeneous multitask
  learning.
\newblock In {\em CVPR}, 2020.

\bibitem{yu2021metaformer}
Weihao Yu, Mi Luo, Pan Zhou, Chenyang Si, Yichen Zhou, Xinchao Wang, Jiashi
  Feng, and Shuicheng Yan.
\newblock {MetaFormer} is actually what you need for vision.
\newblock In {\em CVPR}, 2022.

\bibitem{yuan2020object_contextual_representations}
Yuhui Yuan, Xilin Chen, and Jingdong Wang.
\newblock Object-contextual representations for semantic segmentation.
\newblock In {\em ECCV}, 2020.

\bibitem{domain_randomization}
Xiangyu Yue, Yang Zhang, Sicheng Zhao, Alberto Sangiovanni-Vincentelli, Kurt
  Keutzer, and Boqing Gong.
\newblock Domain randomization and pyramid consistency: Simulation-to-real
  generalization without accessing target domain data.
\newblock In {\em ICCV}, 2019.

\bibitem{wilddash}
Oliver Zendel, Katrin Honauer, Markus Murschitz, Daniel Steininger, and
  Gustavo~Fernandez Dominguez.
\newblock {WildDash} - {Creating} hazard-aware benchmarks.
\newblock In {\em ECCV}, 2018.

\bibitem{zhang2017mixup}
Hongyi Zhang, Moustapha Ciss{\'{e}}, Yann~N. Dauphin, and David Lopez{-}Paz.
\newblock mixup: {Beyond} empirical risk minimization.
\newblock In {\em ICLR}, 2017.

\bibitem{zhang2020resnest}
Hang Zhang, Chongruo Wu, Zhongyue Zhang, Yi Zhu, Zhi Zhang, Haibin Lin, Yue
  Sun, Tong He, Jonas Mueller, R. Manmatha, Mu Li, and Alexander~J. Smola.
\newblock {ResNeSt:} {Split-attention} networks.
\newblock {\em arXiv preprint arXiv:2004.08955}, 2020.

\bibitem{zhang2021generalizable}
Jian Zhang, Lei Qi, Yinghuan Shi, and Yang Gao.
\newblock Generalizable model-agnostic semantic segmentation via
  target-specific normalization.
\newblock {\em PR}, 2022.

\bibitem{zhang2021trans4trans}
Jiaming Zhang, Kailun Yang, Angela Constantinescu, Kunyu Peng, Karin
  M{\"u}ller, and Rainer Stiefelhagen.
\newblock {Trans4Trans:} {Efficient} transformer for transparent object and
  semantic scene segmentation in real-world navigation assistance.
\newblock {\em T-ITS}, 2022.

\bibitem{zhang2022bending}
Jiaming Zhang, Kailun Yang, Chaoxiang Ma, Simon Rei{\ss}, Kunyu Peng, and
  Rainer Stiefelhagen.
\newblock Bending reality: Distortion-aware transformers for adapting to
  panoramic semantic segmentation.
\newblock In {\em CVPR}, 2022.

\bibitem{zhang2020issafe}
Jiaming Zhang, Kailun Yang, and Rainer Stiefelhagen.
\newblock {ISSAFE:} {Improving} semantic segmentation in accidents by fusing
  event-based data.
\newblock In {\em IROS}, 2021.

\bibitem{zhang2021exploring}
Jiaming Zhang, Kailun Yang, and Rainer Stiefelhagen.
\newblock Exploring event-driven dynamic context for accident scene
  segmentation.
\newblock {\em T-ITS}, 2022.

\bibitem{Zhang_2021_CVPR}
Pan Zhang, Bo Zhang, Ting Zhang, Dong Chen, Yong Wang, and Fang Wen.
\newblock Prototypical pseudo label denoising and target structure learning for
  domain adaptive semantic segmentation.
\newblock In {\em CVPR}, 2021.

\bibitem{zhang2017curriculum}
Yang Zhang, Philip David, and Boqing Gong.
\newblock Curriculum domain adaptation for semantic segmentation of urban
  scenes.
\newblock In {\em ICCV}, 2017.

\bibitem{Zhao_2017_CVPR}
Hengshuang Zhao, Jianping Shi, Xiaojuan Qi, Xiaogang Wang, and Jiaya Jia.
\newblock Pyramid scene parsing network.
\newblock In {\em CVPR}, 2017.

\bibitem{zheng2021rethinking}
Sixiao Zheng, Jiachen Lu, Hengshuang Zhao, Xiatian Zhu, Zekun Luo, Yabiao Wang,
  Yanwei Fu, Jianfeng Feng, Tao Xiang, Philip H.~S. Torr, and Li Zhang.
\newblock Rethinking semantic segmentation from a sequence-to-sequence
  perspective with transformers.
\newblock In {\em CVPR}, 2021.

\bibitem{zhou2021context}
Qianyu Zhou, Zhengyang Feng, Qiqi Gu, Jiangmiao Pang, Guangliang Cheng, Xuequan
  Lu, Jianping Shi, and Lizhuang Ma.
\newblock Context-aware mixup for domain adaptive semantic segmentation.
\newblock {\em arXiv preprint arXiv:2108.03557}, 2021.

\bibitem{zhu2019asymmetric}
Zhen Zhu, Mengde Xu, Song Bai, Tengteng Huang, and Xiang Bai.
\newblock Asymmetric non-local neural networks for semantic segmentation.
\newblock In {\em ICCV}, 2019.

\end{thebibliography}
}

\end{document}